\pdfoutput=1

\documentclass[11pt]{article}

\usepackage[]{authblk}
\usepackage{mdframed}
\usepackage{setspace}
\usepackage{multirow}
\usepackage{amsmath} 
\usepackage{float}
\usepackage[table]{xcolor}
\usepackage{multirow}
\usepackage{booktabs}
\usepackage{cjhebrew}
\usepackage{hhline}
\usepackage{CJKutf8}
\usepackage{arydshln}

\usepackage{xspace}

\usepackage{subfig}

\usepackage{graphicx}

\usepackage[T1]{fontenc}

\usepackage{inputenc}
\usepackage{microtype}

\usepackage{inconsolata}
\usepackage{csquotes}

\usepackage{todonotes}

\makeatletter
\renewcommand\AB@affilsepx{, \protect\Affilfont}
\makeatother

\usepackage[]{ACL2023}
\let\oldfootnotetext\footnotetext
\renewcommand{\footnotetext}[1]{%
  \begingroup%
  \renewcommand{\thefootnote}{\ensuremath{*}}%
  \oldfootnotetext{#1}%
  \endgroup%
}

\usepackage{times}

\usepackage{booktabs}
\usepackage{siunitx}
\usepackage{latexsym}
\usepackage{graphicx}
\usepackage{tabu}
\usepackage{multirow}
\usepackage{makecell} 

\usepackage[T1]{fontenc}

\usepackage{microtype}

\usepackage{inconsolata}
\usepackage{csquotes}

\usepackage{todonotes}

\newcommand{\pz}{\hphantom{0}}
\newcommand{\pzz}{\hphantom{00}}

\author[a]{Itai Mondshine}
\author[a]{Tzuf Paz-Argaman}
\author[a]{Reut Tsarfaty}

\affil[a]{Bar-Ilan University, Israel}

\affil[ ]{\authorcr \tt \{mondshi1, tzuf.paz-argaman, reut.tsarfaty\}@biu.ac.il}

\title{Beyond N-Grams: Rethinking Evaluation Metrics and  Strategies
\\ for Multilingual Abstractive Summarization}

\begin{document}

\maketitle
\begin{abstract}

Automatic N-gram based metrics such as ROUGE are widely used for evaluating generative tasks such as summarization. 
While these metrics are considered indicative (even if imperfect), of human evaluation for English, their suitability for other languages remains unclear. To address this, in this paper we systematically assess evaluation metrics for generation --- both n-gram-based and neural-based  --- to assess their effectiveness across languages and tasks. Specifically,  we design a large-scale evaluation suite across eight languages from four typological families --- agglutinative, isolating, low-fusional, and high-fusional --- from  both low- and high-resource languages,  to analyze their correlations with human judgments. 
 Our findings highlight the sensitivity of the evaluation metric to the language type at hand.
For example,
  for fusional languages, n-gram-based metrics demonstrate a lower correlation with human assessments, compared to isolating and agglutinative languages. We also demonstrate that 
tokenization considerations can significantly mitigate this
for fusional languages with rich morphology, up to reversing such negative correlations. Additionally, we show that neural-based metrics specifically trained for evaluation, such as COMET, consistently outperform other neural metrics and correlate better than n-grams metrics with human judgments in low-resource languages. Overall, our analysis highlights the limitations of n-gram metrics for fusional languages and advocates for investment in neural-based metrics trained for evaluation tasks.\footnote {Our human annotation data and the evaluation framework code are publicly available at \url{https://github.com/itaimondshine/Beyond_ngrams}.}



\end{abstract}

\section{Introduction}

The development of multilingual LLMs (MLLMs) such as BLOOM \citep{le2023bloom} and XGLM \citep{lin2021few}, along with the current trend of extending English-centric generative LLMS (e.g.,  OpenAI GPT-4o  \citep{hurst2024gpt}, Gemini 1.5 \citep{team2024gemini} and LLaMA3 \citep{dubey2024llama})
to other languages \citep{alexandrov2024bggpt}, reflects the growing interest in prompting such generative models in languages other than English. This interest highlights the need for robust evaluation of the generation capabilities of LLMs in multilingual settings. However, assessing these models on non-English generative tasks, particularly in summarization, remains challenging due to the lack of clear evaluation methodologies.


Current evaluation metrics for summarization, both n-gram-based and neural-based,  face significant limitations. N-gram-based evaluation metrics, such as BLEU \citep{papineni2002bleu}, ROUGE \citep{lin2004rouge}, and METEOR \citep{banerjee2004meteor}, are commonly used to assess summarization quality in English, however, these metrics rely on complete word units. This creates challenges for fusional languages with flexible word order where inflectional patterns are embedded within word forms.
Moreover, they present difficulties for agglutinative languages, where words have complex internal structures, consisting of multiple morphemes that n-gram-based metrics struggle to capture effectively \citep{abudouwaili2023strategies}. Additionally,
the problem of ambiguity --- where a single form can have multiple meanings --- is amplified in morphologically rich languages (MRLs) as variations in prefixes, suffixes, and root conjugations complicate both comprehension and generation tasks. Moreover, many languages require different tokenization schemes, which poses a challenge for n-gram-based metrics that were originally developed primarily for space-delimited languages, potentially affecting the comparability of evaluations across languages with different scripts and morphological systems. These factors can lead to n-gram-based metrics failing to recognize grammatically correct sentences in generated summaries that convey the intended meaning despite surface-level differences.

Neural network-based approaches for generation evaluation compared against gold references, such as BERTScore \citep{zhang2019bertscore}, depend on the availability of large models trained on large amounts of data and may exhibit poor performance for lower-resourced languages \citep{yousuf2024improving, kaster2021global}. Languages with greater morphological complexity are particularly challenging, as MRLs often produce a large number of infrequent word forms produced by combinations of morphemes, resulting in data sparsity \citep{botev2022deciphering}.




Despite such bits of empirical evidence, while summarization metrics have been extensively studied in English, their applicability to other languages remains understudied. More concretely, existing campaigns for assessing evaluation metrics for generation face three key limitations: (i) {\em lack of language diversity}, resulting in insufficient typological representation---for instance, \citet{koto2021evaluating} excluded languages with high-fusional morphology, and \citet{forde2024re} evaluated only three languages, highlighting scalability concerns; (ii) {\em lack of metrics diversity}, primarily focusing on 
n-gram-based approaches and excluding neural-based ones, particularly those specifically trained for evaluation, and  insufficient evaluation of metric adaptation for non-English; and (iii) {\em lack of reliable statistical evidence} on the correlation between automatic metrics and human judgments, 
omitting statistical significance values of the correlation analysis. \citep{koto2021evaluating, han2024rethinking}


To address these gaps, we deliver a large resource for summarization in non-English languages,  manually annotated with human judgments, comprising \textasciitilde{20,000} human annotations. 
We highlight three upshots of this resources. First, the~\textit{\textbf{selection of representative languages}}, covering eight languages from four typological types (isolating, agglutinative, and languages with minimal or high fusional morphology). Within each group, we represent both high- and low-resource languages.
Second, we assess \textit{\textbf{diverse Metrics}}, both {\em n-gram} and {\em neural-based} metrics, including ones particularly trained for evaluation. Additionally, we evaluate the different methodologies to assess the quality of generation, for example, 
the use of different tokenizers and various transformed versions of the original text, including lemmatized forms, to assess their impact on the evaluation metrics. Finally, our analysis takes care to provide \textit{\textbf{statistically sufficient data size}}. Our multilingual annotation task measures correlation with both n-gram and neural metrics while reporting the statistical significance of the factors found to affect the results.

Our study demonstrates that evaluation metrics perform differently depending on linguistic typology. For instance, n-gram metrics as ROUGE align less reliably with human assessments in fusional languages than in isolating or agglutinative languages. Conversely, neural-based metrics like COMET, trained explicitly for assessing generative task, achieve stronger correlations with human judgments and consistently surpass both n-gram  and neural-based approaches. These findings highlight the limitations of n-gram metrics for fusional languages and emphasize the need for specialized neural metrics trained for multilingual evaluation.





\section{Limitations of Current Generation Evaluation  in Diverse Languages}
\label{chap_2}

\subsection{The Limitations and Shortcomings of Current Generation Evaluation}
The rise of generative large language models (LLMs), and  massive prompting thereof to generate  high-quality online responses, has underscored the importance of properly evaluating such models with automatic metrics  \citep{manduchi2024challenges} that allow effective and efficient hill-climbing in the course of model development and assessment.
Since the introduction of ROUGE \cite{lin2004rouge}, n-gram-based metrics  have been commonly used  for evaluating English tasks as well as for multilingual purposes. However,
these metrics face severe issues with languages that differ from English, specifically those with different tokenization schemes that not align with the common practice of space-delimited metrics. For example, metrics such as BLEU face challenges 
in languages like Chinese and Japanese due to the lack of explicit word boundaries \citep{denoual2005bleu}, and implementations of metrics like ROUGE,
 often struggle with segmentation issues, including filtering out non-alphanumeric Latin characters, making them less effective for non-Latin scripts \citep{kumar2023rouge}. Additionally, these limitations lead to poor correlations with human judgments, especially for high fusional languages. For instance, \citet{bouamor2014human} observed weak correlations for BLEU and METEOR in Arabic, while \citet{paz2024hesum} found negative correlations for ROUGE in Hebrew.

To address the limitations of n-gram-based metrics, researchers proposed to utilize neural-based metrics,
which fall into three categories: \textit{encoder-based models} like BERTScore~\cite{zhang2019bertscore}, which compare text representations; \textit{LLM-as-a-judge} methods, such as the prompting of Gemini~\cite{team2023gemini} to assess quality without any parameter updates; and \textit{neural methods specifically trained for evaluating generation}
such as COMET~\cite{rei2020comet}, fine-tuned to predict quality scores for machine translation (MT).
These metrics, while remaining data-driven and agnostic to the language type at hand, are prone to suffer from resource-level effects with varying qualities that depend on the model exposure to such data.
All in all, both  {\em n-gram}  and {\em neural based} metrics (including those specifically trained for evaluation) have not been  systematically evaluated for non-English languages.
To the best of our knowledge, this  is the first work to provide a systematic multilingual assessment of  metrics for generation.

\subsection{Generation Evaluation in the Face of Language Diversity}
\label{language_families}
Despite their shortcomings, the effectiveness of n-gram-based as well as neural based metrics for evaluation of generation has not been systematically studied across language families with varying word complexity and boundary characteristics. 
This raises concerns, as the linguistic properties of words may well affect the usability of n-gram metrics,  
but the effects remain unclear. 
Let's elaborate.

In terms of their linguistic properties, language families can be placed on a scale. On the one hand, there are \textbf{Isolating Languages}, in which words typically consist of a single morpheme, e.g., Yoruba and Chinese \citep{okanlawon2016analysis, arcodia2007chinese}. On the other hand, words in \textbf{Fusional Languages} contain multiple morphemes fused together, often with unclear boundaries, where a single space-delimited token may serve multiple functions. For example, in the Spanish word \textit{habló}, the suffix \texttt{ó} simultaneously indicates past tense and third-person singular \citep{kambarami2021computational}. This category can be further divided into \textbf{low-fusional} (e.g. Spanish \citep{bergmann2007language} and Ukrainian \citep{budzhak1998against}) and \textbf{high-fusional} (e.g. Arabic \citep{smrvz2007functional} and Hebrew \citep{tsarfaty2019s})  based on the degree of morphological fusion. Additionally,
in an orthogonal dimension we can recognize \textbf{Agglutinative Languages} that also consist of words made up of multiple morphemes, albeit with clear boundaries and distinct functions. For instance, in Shona, \textit{vakaenda} (\texttt{va-ka-end-a}) means “they went” where \texttt{va} (plural subject), \texttt{ka} (remote past), and \texttt{a} (final vowel) modify the root \texttt{end} (“to go”) \citep{kambarami2021computational}. Examples include Turkish and Japanese \citep{istek2007link, shibatani2015introduction}. To our knowledge, no non-English evaluation has comprehensively covered languages from all these typological groups.

Two primary strategies have been suggested to adapt previously used metrics to different types of languages. First, for instance, is \textit{data transformation},    the adaptation of n-gram metrics, where a different tokenizer or lemmatizer is applied to the data prior to  using the n-gram-based metrics. Specifically, converting Chinese text into numerical IDs before applying ROUGE \citep{wang2021cnewsum}, or using ROUGE with language-specific tokenizers as \citet{alhamadani2022lans} did for Arabic. Alternatively, researchers suggested the use of  \textit{language-specific encoders}, encoders trained on the target language for similarity-based evaluation against a gold reference text. For example,  using BERTScore with language-specific models \citep{vetrov2022new}. However,  these approaches have not been systematically evaluated across languages.


In addition to the lack of coverage in languages and metrics,
correlations between multilingual automatic metrics and human judgments lack sufficient evidence to be considered reliable due to the absence of reported p-values \citep{koto2021evaluating, forde2024re, han2024rethinking}. In reproduced experiments \citep{ernst-etal-2023-examining}, the statistical significance was low 
to substantiate the findings. Additionally, power analysis indicates that \textasciitilde{400} samples per language are needed to detect significant effects at \( p \leq 0.05 \).\footnote{See Appendix \ref{appendix:power_test} for more details on the t-test.} However, existing non-English evaluations fall short of this threshold, with \citet{koto2021evaluating} using  150 samples and \citet{han2024rethinking} evaluating  90 summaries per language.

\section{Our Approach: Systematic Evaluation of  Summarization Across Languages}



In this work we set out  to systematically evaluate automatic metrics for text generation, assessing their effectiveness and reliability for non-English languages by assessing the correlation 
with human scores. 
We do so via a comprehensive and controlled protocol, comprising \textasciitilde{20,000} human annotations while addressing the various  diversity dimensions and previously attested weaknesses. 

Concretely, in this work 
we evaluate eight languages from four typological families, covering both low resource (L) and high resource (H) language in each group, including: \textit{Isolating} (Chinese, \textit{zh} (H); Yoruba, \textit{yo} (L)), \textit{Agglutinative} (Japanese, \textit{ja} (H); Turkish, \textit{tr} (L)), \textit{Low Fusional} (Spanish, \textit{es} (H) and Ukrainian, \textit{ukr} (L)) and \textit{High Fusional} (Arabic, \textit{ar} (H); Hebrew, \textit{he} (L)).
We followed \citet{lai2023chatgpt}'s  method in classifying H/L languages using a threshold, and classified languages by token percentage based on GPT-3's pre-trained data distribution,
relying on its broad multilingual coverage and reported data mix.\footnote{\url{https://github.com/openai/gpt-3/blob/master/dataset_statistics}} Specifically, we classified languages into low- (\(< 0.1\%\)) and high-resource (\(\geq 0.1\%\)).\footnote{Arabic, with less than 0.1\% of tokens, was chosen as a high-resource language due its worker availability and higher pre-trained representation than Hebrew. See Appendix \ref{appendix:language_classes} for the full language proportions.}  For language selection within each typological family, we followed \citet{gerz2018language} (see Section \ref{language_families} for additional justifications).


For each language-metric combination
we perform a correlation analysis with both general purpose metrics, as well as metrics tailored for multilingual settings, e.g., BERTScore applied with mBERT, or with BERT models trained monolingually. Also, we have utilized COMET \citep{rei2020comet} --- a neural framework for machine translation evaluation with a model trained multilingually. Additionally, we have used the ROUGE score with different definitions of wordhood.\footnote{See Appendix \ref{appendix:multilingual_metrics} for all models and tokenizers we used.} Finally, to substantiate our results, we included at least 400 samples per language and reported p-values
for each evaluated dimension. For all experiments, we report inter-annotator agreement to assess the credibility of our annotations.

\section{Data Collection}
\label{chap_3}

To systematically assess the correlation between evaluation metrics and human rankings for abstractive summarization, we engage human annotators to evaluate summaries generated by large language models (LLMs). Our data collection evaluates document summaries in eight languages, chosen to represent four typological families with both low- and high-resource languages within each group. The annotators rank the summaries along two quality dimensions: \textit{coherence}, which assesses the summaries’ grammaticality and readability, and \textit{completeness}, which measures the degree to which they capture the text's main ideas.







\begin{table}[t]
\resizebox{\columnwidth}{!}{%
\begin{tabular}{lcccc}
\hline
\textbf{Resource/Type}                            & \textbf{Isolating}               & \textbf{Agglutinative} & \textbf{High Fusion} & \textbf{Low Fusion}   \\ \hline
\multicolumn{1}{l|}{\textbf{High Resource}} & Simplified Chinese (\textit{zh}) & Japanese (\textit{jp}) & Arabic (\textit{ar}) & Spanish (\textit{es}) \\
\multicolumn{1}{l|}{\textbf{Low Resource}}  & Yoruba (\textit{yor})             & Turkish (\textit{tr})  & Hebrew (\textit{he}) & Ukraine (\textit{ukr}) \\ \hline
\end{tabular}%
}
\caption{Categorization of languages based on morphological typology and resource availability. ISO 639-1 language codes are provided in parentheses.}
\label{tab:language_categorization}
\end{table}

\begin{table*}[t]
\centering
\scalebox{0.77}{
\begin{tabular}{lccccccccc}
\toprule
Family & Language (L/H) & \multicolumn{4}{c}{Novel n-grams} & \multicolumn{2}{c}{Redundancy} & Compression & Mean Token Length \\
\cmidrule(lr){3-6} \cmidrule(lr){7-8}
 & & 1-gram & 2-gram & 3-gram & 4-gram & n=1 & n=2 & & \\
\midrule
Isolating & ZH (H)       & 27.52  & 67.23  & 83.82  & 91.29  & 14.86 & 2.34 & 83.71 & 53.56  \\
                 & YOR (L)       & 38.90  & 60.85  & 69.38  & 73.84  & 32.85 & 8.03 & 62.17 & 105.29 \\
\midrule
Agglutinative & JP (H)        & 24.29  & 54.12  & 69.62  & 78.23  & 49.08 & 15.93 & 79.22 & 188.37 \\
                    & TR (L)       & 41.76  & 71.44  & 84.56  & 90.76  & 18.41 & 2.37 & 72.71 & 69.95  \\
\midrule
Low Fusional & ES (H)        & 28.00  & 63.15  & 81.16  & 89.11  & 26.28 & 2.83 & 81.94 & 83.17  \\
                    & UKR (L)       & 42.01  & 73.49  & 86.72  & 92.39  & 18.53 & 2.21 & 74.85 & 66.22  \\
\midrule
High Fusional & AR (H)        & 47.73  & 78.72  & 89.75  & 94.59  & 15.05 & 1.62 & 77.36 & 62.32  \\
                      & HE (L)       & 45.06  & 75.14  & 86.75  & 92.01  & 20.83 & 3.49 & 84.28 & 80.85  \\
\bottomrule
\end{tabular}}
\caption{Model-Generated Summaries Intrinsic Evaluation per language.
}
\label{tab:abstrctness}
\end{table*}

\begin{table}[t]
\centering
\scalebox{0.6}{
\begin{tabular}{lcc}
\toprule
\textbf{Country of Residence} & \textbf{Total Workers} & \textbf{Percentage (\%)} \\
\midrule
United States                & 5  & 13.9 \\
Nigeria                      & 2  & 5.6  \\
West Africa                  & 2  & 5.6  \\
Turkey                       & 3  & 8.3  \\
Egypt                        & 1  & 2.8  \\
Jordan                       & 1  & 2.8  \\
mibya                        & 2  & 5.6  \\
Ukraine                      & 5  & 13.9 \\
Israel                       & 5  & 13.9 \\
Spain                        & 4  & 11.1 \\
Mexico                       & 1  & 2.8  \\
Argentina                    & 2  & 5.6  \\
Venezuela                    & 2  & 5.6  \\
Japan                        & 1  & 2.8  \\
\midrule
\textbf{Total}               & 36 & 100.0 \\
\bottomrule
\end{tabular}
}
\caption{\textit{Distribution of Workers by Country of Birth.}}
\label{tab:workers}
\end{table}


\subsection{The Generated Summaries}
We used the XL-Sum dataset \citep{hasan2021xl}, which provides news articles along with their human-generated summaries in various languages. For Hebrew, we used HeSum \citep{paz2024hesum}. See Table \ref{tab:language_categorization} for categorization details. 

First, we generated two parallel summaries—produced by GPT-3.5-Turbo (0125) (Ouyang et al., 2022) and Gemini 1.0 Pro (Team et al., 2023) on 400 random samples from each language's test split.

Secondly, to achieve a diverse distribution of scores, we artificially corrupted one-third of the data by randomly degrading one quality criterion.\footnote{We adopted this approach following a previous data collection experiment without  such  corruption, which revealed scores that were too clustered and displayed low dispersion.} 
For coherence, we replaced nouns and verbs with their lemma forms, creating ungrammatical sentences. Additionally, we reordered non-adjacent sentences to disrupt the flow. For completeness, we replaced named entities in the summary with others from the original text and inserted a random, unrelated sentence.
\footnote{See Appendix \ref{appendix:corruption} for  complete details on the corruption.}

\subsection{The Task: Ranking the  Generated Summaries }
The task involves annotating two parallel summaries by comparing their content to the source article. The evaluation procedure is as follows: \textit{(i)}~The annotator reads the source article and the two summaries. \textit{(ii)}~The annotator answers a question on the article to prove language comprehension.
\textit{(iii)}~The annotator evaluates each summary using 1-4 Likert scale \cite{likert1932technique} based on two quality criteria (QC): \textit{coherence}, and \textit{completeness}. The evaluation page was set up to include the full source article, instructions, definitions of the quality criteria, and two generated summaries. For each summary and criterion, there is a scale with four rating options. Appendix \ref{appendix:ui} presents the UI interface we designed and built for the assignment as displayed to the annotators in Arabic and Spanish. Appendix \ref{appendix:protocol_details} gives more details about the collection protocol. 


\subsection{Ensuring High Annotation Consistency}

To ensure annotation reliability, we hired annotators through Amazon Mechanical Turk (MTurk) (100+ approved HITs, 90\%+ approval rate) with geographic constraints aligned to the target languages. 
For Yoruba and Japanese, we were unable to recruit native speakers in their country of birth due to various restrictions and sourcing difficulties; in such cases, we hired native speakers residing in other countries.\footnote{In these cases, we used the qualification question to assess the participant's language skills.} Additionally, we recruited qualified students who passed a matching questionnaire. In total, we recruited 36 raters across 13 locales.\footnote{See Table \ref{tab:workers} for participants' demographics.}
To improve annotation quality, each model-generated summary was ranked by three different participants. For correlation analysis, we used the average score. 

To verify understanding of the source content, we created a Gemini-generated qualification question based on the article to filter annotations from disqualified workers.\footnote{See Appendix \ref{appendix:qualification_task} for  details on the qualification task.} To measure the consistency of the
annotators’ scores, we calculated for each language the Krippendorff’s $\alpha$ \citep{krippendorff2011computing} for an interval scale. 


\section{
Correlation Analysis Settings}

Based on the collected data, including both the generated summaries and human annotations, we present our data analysis in Section \ref{data_analysis}. The complete list of evaluation metrics used is detailed in Section \ref{metrics}. 


\subsection{Data Analysis}
\label{data_analysis}


\paragraph{Generated Summaries Analysis}
To empirically quantify the properties of the model-generated summaries we use 4 established metrics: 
(i) \emph{Abstactness (novel n-grams)} -- 
  the percentage of summary n-grams absent in the article \citep{narayan2018don}.
(ii) \emph{Redundancy (RED)}  -- 
measures repetitive n-grams within a summary (S) using the formula:   \( RED(S) = \frac{{\sum_{i=1}^{m} (f_i - 1)}}{{\sum_{i=1}^{m} f_i}}\) where  \( m \) is the number of unique n-grams in the summary and \( f_i \) represents a frequency of specific n-gram
within the summary.
(iii) \emph{Compression Ratio (CMP)} -- 
 the word counts in summary (S) divided by the corresponding article (A): \( CMP_w(S, A) = 1 - \frac{{|S|}}{{|A|}} \). 
Higher compression ratios result in greater reduction at the word level, which can make the summarization task more difficult \citep{bommasani2020intrinsic}. (iv) \emph{Mean Token Length} -- The average token count per summary by a word-delimited tokenizer. 

Table~\ref{tab:abstrctness} presents a quantitative analysis of the characteristics of model-generated summaries, highlighting the challenges in evaluating our data. We hypothesize that languages with a high level of abstractness (>35 novel 1-grams) are more difficult to evaluate using n-gram-based metrics, which rely on overlap matching, due to their novel, distilled, and non-redundant nature. This challenge is particularly pronounced in high-fusion languages, which often exhibit more complex linguistic structures in addition to their abstractness.

\paragraph{Human Annotations Analysis}
Table~\ref{tab:data_statistics} presents the statistics of the collected human annotations across languages. The average agreement rate, measured using Krippendorff’s $\alpha$, is 0.4 for coherence and 0.47 for completeness, indicating moderate inter-annotator agreement. 
In Table~\ref{tab:data_statistics}, we observe that the mean absolute gap between the scores assigned to Gemini- and GPT-generated summaries is $\sim1$ across all languages, for both coherence and completeness. 
This gap demonstrates the effectiveness of the applied corruption in diversifying the quality of the summaries.
Additionally, the data analysis helps identify languages with higher levels of human disagreement on the generated summaries. We hypothesize that languages with a low agreement rate (e.g., Arabic) will exhibit weaker correlations with automatic metrics, while those with high agreement rates (e.g., Japanese) will show stronger correlations.

\begin{figure}[t]
\centering
\includegraphics[width=0.48\textwidth]{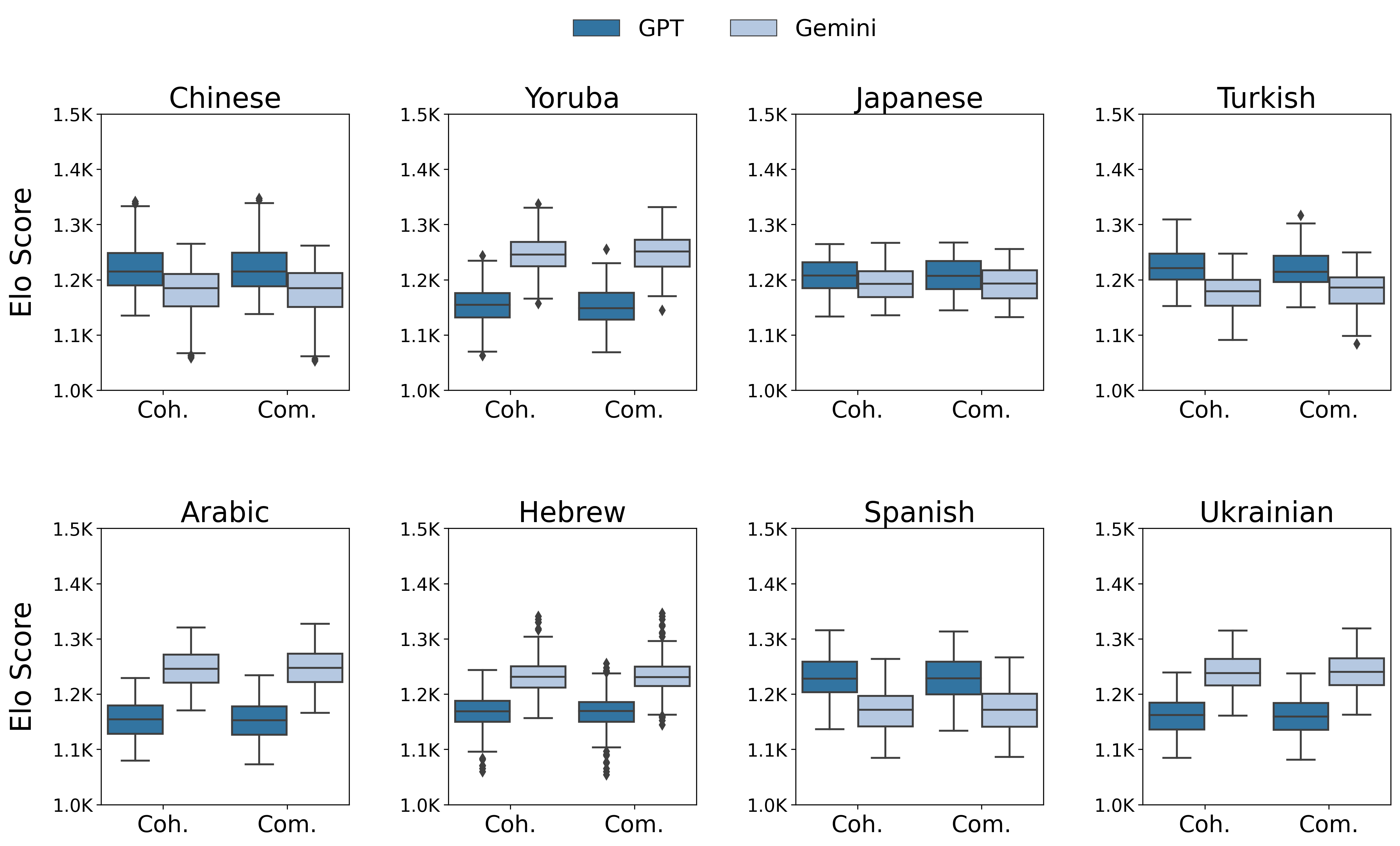}
\caption{Elo score distribution of human annotations for Gemini- and GPT-generated summaries across all criteria. Coh. = Coherence, Com. = Completeness.}
\label{fig:elo_scores}%
\end{figure}

Additionally, we used Elo rankings \citep{elo1978rating} to compare the performance of the two models (Gemini and GPT, including the manually corrupted summaries). Following the implementation of \citet{gong2024cream}, we treat each pairwise human annotation as a comparison between the two models, where each model is represented by its generated summary. After each comparison, we update the models’ Elo scores: the model whose summary is preferred gains points, while the other loses points. This iterative process, based on the standard Elo update rule, yields a relative ranking of the models for each quality criterion and language, as shown in Figure \ref{fig:elo_scores}.
For all languages, the best-performing model is ranked higher on both criteria, which may be attributed to the \textit{halo effect}, where an overall positive impression influences judgments across multiple aspects \citep{draws2021checklist}.
Interestingly, we observe that summaries generated by Gemini (overall with and without corruption) generally rank higher for high-fusional and low-resource languages, while GPT summaries (with and without corruption) are ranked higher for high-resource languages.

\begin{table}[t]
\centering
\scalebox{0.62}{
\begin{tabular}{lccccccc}
\toprule
\textbf{Lang.} & \multicolumn{2}{c}{\textbf{Agreement}} & \multicolumn{2}{c}{\textbf{Avg. Score (Std)}} & \multicolumn{2}{c}{\textbf{Avg. Gap (Std)}} & \textbf{\# Ann.} \\
\cmidrule(lr){2-3} \cmidrule(lr){4-5}  \cmidrule(lr){6-7}
 & \textbf{Coh.} & \textbf{Com.} & \textbf{Coh.} & \textbf{Com.} & \textbf{Coh.} & \textbf{Com.} & \\
\midrule
ZH & 0.35 & 0.35 & 3.2 (0.8) & 3.2 (0.8) & 1.0 (0.7) & 1.0 (0.8) & 1504 \\
YOR & 0.40 & 0.49 & 3.0 (0.9) & 3.1 (0.8) & 1.0 (0.8) & 0.9 (0.7) & 1296 \\
JA & 0.61 & 0.40 & 3.5 (0.7) & 3.4 (0.7) & 0.8 (0.8) & 0.7 (0.6) & 188 \\
TR & 0.32 & 0.40 & 3.2 (0.9) & 2.9 (1.0) & 1.0 (0.9) & 1.3 (0.9) & 2200 \\
AR & 0.32 & 0.35 & 2.6 (0.8) & 2.7 (0.7) & 0.8 (0.8) & 0.9 (0.7) & 1352 \\
HE & 0.71 & 0.65 & 3.8 (1.1) & 3.5 (1.2) & 0.9 (0.9) & 0.9 (0.9) & 1284 \\
ES & 0.42 & 0.42 & 3.2 (0.9) & 3.1 (0.7) & 1.0 (1.0) & 0.7 (0.7) & 1464 \\
UKR & 0.46 & 0.62 & 3.3 (0.8) & 3.2 (0.8) & 0.8 (0.9) & 0.9 (0.8) & 2212 \\
\bottomrule
\end{tabular}

}
\caption{\textit{Human Annotation Statistics:} Krippendorff’s $\alpha$ (agreement), average score, mean absolute gap between Gemini and GPT annotations, and annotation count per language. Coh. = Coherence, Com. = Completeness.}
\label{tab:data_statistics}
\end{table}

\begin{table*}[t]
\centering
\scalebox{0.57}{
\begin{tabular}{lcccccccccc}
\toprule
\textbf{Criteria} & \multicolumn{4}{c}{\textbf{Coherence}} & & \multicolumn{4}{c}{\textbf{Completeness}} \\
\cmidrule{2-5} \cmidrule{7-10}
\textbf{Typological Family} & \textbf{Isolating} & \textbf{Agglutinative} & \textbf{Low Fusional} & \textbf{High Fusional} & & \textbf{Isolating} & \textbf{Agglutinative} & \textbf{Low Fusional} & \textbf{High Fusional} \\
\midrule
\rowcolor{gray!12} 
\multicolumn{10}{c}{\textbf{N-Gram Metrics}} \\ 
\midrule
\makebox[0pt][r]{\scriptsize 1}\hspace{1em}ROUGE1 & \textbf{0.20}** & \pz\pz0.27** & 0.11* & \pz\pz-0.25** & & \pz\pz0.15** & \pz0.11** & 0.08* & \pz\pz-0.20** \\
\makebox[0pt][r]{\scriptsize 2}\hspace{1em}ROUGE2 & 0.20** & \pz\pz0.28** & 0.11* & \pz\pz-0.07** & & \pz\pz0.14** & \pz0.14** & 0.08* & -0.03 \\
\makebox[0pt][r]{\scriptsize 3}\hspace{1em}ROUGE3 & 0.16** & \pz\pz0.27** & 0.09* & \pz\pz-0.01** & & \pz\pz0.12** & 0.10* & 0.01* & \pz0.02 \\
\makebox[0pt][r]{\scriptsize 5}\hspace{1em}ROUGEL & 0.19** & \pz\pz0.23** & 0.11* & \pz\pz-0.23** & & \pz\pz0.15** & 0.10* & 0.08* & \pz\pz-0.18** \\
\makebox[0pt][r]{\scriptsize 6}\hspace{1em}BLEU & 0.03** & \pz0.03 & \pz0.11** & \pz\pz-0.30** & & 0.02 & 0.05* & 0.07* & \pz\pz-0.10** \\
\makebox[0pt][r]{\scriptsize 7}\hspace{1em}CHRF & 0.02** & 0.09 & \pz0.16** & \pz\pz-0.46** & & \pz0.01* & 0.01* & 0.14* & \pz\pz-0.38** \\
\hdashline
\makebox[0pt][r]{\scriptsize 8}\hspace{1em}ROUGE1 (mBERT Tokenizer) & 0.14** & \pz\pz0.18** & \pz0.15** & \pz\pz0.10** & & \pz0.10* & 0.09* & \pz0.14** & \pz\pz0.15** \\
\makebox[0pt][r]{\scriptsize 9}\hspace{1em}ROUGE2 (mBERT Tokenizer) & 0.14** & \pz\pz0.20** & \pz0.15** & \pz0.11* & & \pz0.10* & 0.09* & \pz0.19** & \pz\pz0.15** \\
\makebox[0pt][r]{\scriptsize 10}\hspace{1em}ROUGE3 (mBERT Tokenizer) & 0.12** & \pz\pz0.22** & \pz0.12** &\pz0.11* & & \pz0.10* & 0.07* & \pz0.15** & \pz\pz0.14** \\
\makebox[0pt][r]{\scriptsize 11}\hspace{1em}ROUGEL (mBERT Tokenizer) & 0.14** & \pz\pz0.17** & \pz0.13** & \pz0.08* & & \pz0.11* & 0.05* & \pz0.13** & \pz\pz0.12** \\
\makebox[0pt][r]{\scriptsize 12}\hspace{1em}ROUGE1 (Monolingual) & 0.17** & \pz\pz0.23** & \pz0.11** & \pz0.02* & & 0.07 & 0.13* & 0.06* & \pz\pz0.07** \\
\makebox[0pt][r]{\scriptsize 13}\hspace{1em}ROUGE2 (Monolingual) & 0.12** & \pz\pz0.25** & \pz0.12** & 0.09 & & \pz0.12* & 0.13* & 0.07* & \pz\pz0.14** \\
\makebox[0pt][r]{\scriptsize 14}\hspace{1em}ROUGE3 (Monolingual) & 0.07** & \pz\pz0.24** & \pz0.13** & 0.07 & & 0.07 & 0.08* & 0.02* & \pz\pz0.09* \\
\makebox[0pt][r]{\scriptsize 15}\hspace{1em}ROUGEL (Monolingual) & 0.10** & \pz\pz0.22** & \pz0.11** & 0.03 & & 0.08 & 0.12* & 0.07* & \pz\pz0.11* \\
\makebox[0pt][r]{\scriptsize 16}\hspace{1em}BLEU (Lemmatized Form) & N.A & N.A & 0.15** & \pz\pz0.30** & & N.A & N.A* & \pz0.08* & \pz\pz\textbf{0.40}* \\
\midrule
\rowcolor{gray!12} 
\multicolumn{10}{c}{\textbf{Neural-Based Metrics}} \\ 
\midrule
\makebox[0pt][r]{\scriptsize 17}\hspace{1em} Gemini as a Judge & 0.15** & 0.03 & 0.15* & \pz0.05* & & \pz\pz0.14** & \pz0.16** & \pz0.10** & \pz\pz0.09** \\
\midrule
\makebox[0pt][r]{\scriptsize 18}\hspace{1em} MoverScore & 0.07** & \pz0.15* & 0.18* & 0.02 & & 0.08 & 0.10* & \pz0.17** & \pz0.08* \\
\makebox[0pt][r]{\scriptsize 19}\hspace{1em} BERTScore (mBERT) & 0.09** & \pz\pz0.15** & \pz0.19** & \pz0.15* & & \pz\pz0.13** & 0.07* & \pz0.16** & \pz\pz0.13** \\
\makebox[0pt][r]{\scriptsize 20}\hspace{1em} BERTScore (Monolingual) & 0.13** & \pz\pz\textbf{0.32}** & \pz0.20** & \pz\pz0.17* & & \pz\pz0.12** & \pz\textbf{0.21}** & -0.03* & \pz\pz0.15** \\
\midrule
\makebox[0pt][r]{\scriptsize 21}\hspace{1em} COMET & 0.07** & \pz\pz0.23** & \pz\textbf{0.23}** & \pz\textbf{0.35}* & & \pz\pz\textbf{0.16}** & \pz0.18** & \pz\textbf{0.24}** & \pz\pz0.24** \\
\bottomrule
\end{tabular}
}
\caption{Pearson correlation between resource types and evaluation metrics. Significance: * $p < 0.05$, ** $p < 0.01$. The dashed line separates English-based from multilingual metrics. The highest correlation per column is in bold.}
\label{tab:family_correlations}
\end{table*}

\subsection{Assessed Metrics for Summarization}
\label{metrics}


We assess a total of 10
evaluation metrics that are common in evaluating abstractive summarization: 

\textbf{N-Gram Metrics:} measure the lexical overlap between the system and reference summaries. For this evaluation, we used \textit{\textbf{ROUGE}} \citep{lin2004rouge}, considering four variants: ROUGE-1 (unigram), ROUGE-2 (bigram), ROUGE-3 (trigram), ROUGE-L (longest common subsequence).
We also use \textit{\textbf{CHRF}} \citep{popovic2015chrf}  measuring the character n-gram F-score; and \textit{\textbf{BLEU}} \citep{papineni2002bleu}.
We also utilized  n-grams metrics adapted for multilingual use by means of pre-tokenization: \textbf{\textit{ROUGE + an mBERT Tokenizer}} leverages Byte-Pair Encoding (BPE) tokenization from BERT-multilingual \citep{kenton2019bert}, 
and \textbf{\textit{ROUGE  
 + a Monolingual tokenizer}} is  equipped with a language-specific tokenizer enabling the adaptability to specific languages.\footnote{See Appendix \ref{appendix:multilingual_metrics} for the full list of the tokenizers  used.}

\textbf{Neural-Based Metrics:} \textit{\textbf{MoverScore}} \citep{zhao2019moverscore}  measures the Euclidean distance between two contextualized BERT representations of the paragraphs and finds an optimal soft alignment through an optimization process. We utilized this metric with mBERT to support adaptation across all languages.
BERTScore \citep{zhang2019bertscore} computes the similarity between BERT token embeddings of the system and reference summaries. For multilingual evaluation, we  used two variants: \textbf{\textit{BERTScore (mBERT)}} which was trained on 104 languages \citep{kenton2019bert}, and \textbf{\textit{BERTScore (Monolingual)}} based on a language-specific BERT model.
We also used  \textbf{\textit{Gemini as a Judge}}  \citep{team2023gemini} ---  with the Gemini model 1.0-pro as an evaluator, in which the given prompt was in the same format as the one given to the annotators. Finally, we utilized \textit{\textbf{COMET}} \citep{rei2020comet}, a framework for machine translation evaluation (MT) using a regression-based objective to minimize the mean squared error (MSE) between predicted quality scores and human-annotated scores.  
Specifically, we used the 
pre-trained model \emph{wmt22-comet-da}, built on the XLM-R model \citep{conneau2019unsupervised} and trained for machine translation evaluation as mentioned above.
We adapt COMET for summarization evaluation by excluding the source input, as summarization assessment focuses on comparing the generated summary to a human-written reference. While COMET has been designed exclusively for MT, we extended its applicability to summarization evaluation, as both tasks involve evaluating a generated output against a gold reference. To the best of our knowledge, this is the first work to suggest COMET for evaluating a non-MT task.

\section{Results and Analysis}
\label{results}

\paragraph{Goal} Based on the human annotations of generated summaries, we are now ready to examine the Pearson correlation between the human annotations with both n-gram and neural metrics. We aim to investigate what influences the correlation and to  systematically assess the ways that have been proposed to mitigate poor correlations.
To achieve this, we analyze several aspects, 
including language typology family, resource availability, and metrics that are adapted to multilingual evaluation.  
Table~\ref{tab:family_correlations} shows the correlation from a language type perspective, while Table~\ref{tab:correlations_neural} presents the correlation from a resource-type perspective. 
See the Appendix \ref{appendix:correlation_results_apendix} for the particular correlations per language.\footnote{Also for correlations  using Spearman's rank correlation.}



\paragraph{The Impact of Typological Family}

Table \ref{tab:family_correlations} examines the Pearson correlations from the typological family perspective. The correlations for each family were measured across all the languages within the respective linguistic family. Overall, it appears that n-gram metrics are sensitive to the typological family of the language, while neural metrics have not shown this tendency. For example, for both criteria, fusional languages exhibit weaker correlations with human judgments, with low correlations for Low-Fusional languages and even negative correlations for High-Fusional languages, due to their rich morphology (lines 1-7). However, for neural-based metrics, the typological family appears to play a less critical role. For instance, low-fusional languages achieve the highest correlation for BERTScore (mBERT) (line 19) in both criteria. 
Interestingly, COMET exhibits an inverse trend compared to n-gram metrics, consistently showing a better correlation with fusional languages (line 21). Additionally, the results for n-gram metrics not adapted to multilingual settings (lines 1-7) show that agglutinative languages displayed better correlations with human scores than 
Isolating languages in coherence, while Isolating languages show a better correlation in completeness.
The advantage of agglutinative languages over Isolating languages is surprising, given that these families tend to have more complex morphological structures due to longer morphemes, which may be more challenging for tokenizers.\footnote{We acknowledge that the disparity may stem from the poor quality of generated summaries in Yoruba, a low-resource language compared to Turkish. We hypothesize that the low generation quality contributed to the weak performance of automatic metrics, despite the relatively high human scores in Table \ref{tab:data_statistics}, which may explain the low correlation observed.} Overall, neural-based metrics show a stronger correlation than n-gram-based metrics.

\paragraph{The Impact of Resource Level}

Table \ref{tab:correlations_neural} presents the Pearson correlation between human annotations (by resource type) and neural metrics, evaluating coherence and completeness for high- and low-resource languages.
The results indicate that Gemini-as-a-judge exhibits the lowest correlations with human scores among other multilingual neural metrics for both criteria, regardless of language resource level, indicating that LLMs as judges still lag behind other metrics. Furthermore, the table presents an advantage for language-specific BERT models over multilingual BERT, suggesting that a dedicated monolingual model improves correlation more than training on larger, non-specific datasets maybe due to the data size it was trained on.\footnote{A comprehensive list of the BERT models employed in this study is provided in Appendix \ref{appendix:multilingual_metrics}.}  


Notably, COMET shows the strongest correlation with human scores for coherence across both high- and low-resource types, and for completeness in the low-resource setting. This can be attributed to COMET’s targetted training of  evaluation for generative tasks, enabling it to better capture human-like evaluation. This is particularly useful  in challenging scenarios such as low resource setting. Its performance underscores the potential of task-specific training to bridge the gap between automated metrics and human evaluation, particularly for low-resource languages. We hypothesize that a metric trained specifically for summarization evaluation could perform even better.


\begin{table}[t]
\centering
\scalebox{0.67}{
\begin{tabular}{lcccc}
\toprule
\textbf{Criteria} & \multicolumn{2}{c}{\textbf{Coherence}} & \multicolumn{2}{c}{\textbf{Completeness}} \\
\cmidrule{2-3} \cmidrule{4-5}
\textbf{Resource Type} & \textbf{High} & \textbf{Low} & \textbf{High} & \textbf{Low} \\
\midrule
\midrule
Gemini as a Judge & 0.19* & 0.13** & \pz\pz0.08** & \pz\pz0.12** \\
\midrule
MoverScore & \pz0.16** & 0.13** & \pz0.10* & \pz0.06*\\
BERTScore (mBERT)   & \pz0.23** & 0.16** & \pz\pz0.16** & \pz\pz0.15** \\
BERTScore (Monolingual) & \pz0.27** & 0.16** & \pz\pz0.17** & \pz\pz0.21** \\
\midrule
COMET & \pz0.32** & 0.18** & \pz\pz0.13** & \pz\pz0.24** \\

\bottomrule
\end{tabular}
}
\caption{Pearson correlation between low- and high-resource human annotations and neural-based metrics. significance levels denoted by: * \(p < 0.05\), ** \(p < 0.01\).}
\label{tab:correlations_neural}
\end{table}


\vspace{-2pt}


\paragraph{The Impact of Metrics Adapted to non-English Languages}

The results in Table~\ref{tab:family_correlations} highlight the importance of adequate tokenizers for fusional languages and in particular for isolating and agglutinative languages in completeness evaluation (lines 1-7 vs. 8-15). For example, ROUGE with mBERT tokenizer or a language-specific tokenizer (lines 8–15) improves correlation and can even reverse a negative correlation to a positive one in languages with highly morphological grammar, such as Hebrew and Arabic (e.g., ROUGE-L in high-fusional languages improves from -0.23 to 0.08, lines 5 \& 11). 
Also, applying BLEU to the lemmatized text shows a significant improvement for fusional languages, with the correlation increasing from -0.10 to 0.40 for high-fusional languages (line 6 vs. 16).

Notably, for isolating and agglutinative, correlations decrease, favoring the space-delimited ROUGE variation. We hypothesize that tokenizers struggle with the long morphological sequences in agglutinative languages, making it difficult to split morphemes correctly. As a result, tokenization with space delimitation may be more effective. However, for completeness, the adapted variations have shown better performance. The inverse correlation is also observed, with positive correlations for BERTScore variations and MoverScore in high-fusional languages (lines 18-20). Additionally, using models not trained on non-English languages is suboptimal, as shown in Table~\ref{tab:correlations_neural}, where MoverScore—untrained on non-English—performs worst for both coherence and completeness.



\section{Conclusion}

In this work, we systematically evaluate the reliability of automatic metrics of evaluation for text generation in non-English languages, through a comprehensive correlation analysis with human annotations. We aim to identify the  factors that influence these correlations and asses new metrics variants and approaches designed for the multilingual summarization evaluation task.



Our annotation protocol addresses previous weaknesses, including limited typological family and resource type coverage, insufficient evaluation of diverse metrics (particularly neural-network-based models trained for evaluation), and adaptation of general-purpose metrics to non-English languages. Also,  unlike prior non-English evaluations, we provide statistical significance reports of the results. We crowd-sourced rank annotations for eight languages representing diverse typological families, each with different word boundaries, a key factor for n-gram-based metrics.  We further included both high- and low-resource languages within each typological group, as resource levels potentially affect the reliability of neural metrics.



Our analyses highlight the limited ability of n-gram-based metrics to handle complex linguistic structures—particularly those found in fusional languages—compared to neural network-based metrics, especially ones trained for multilingual evaluation of generative models. Based on these findings, we recommend transitioning from n-gram metrics to neural models specifically trained for multilingual summarization. As a possible mitigation during this transition, when using n-gram metrics for fusional languages, we suggest employing tokenization techniques that break down complex linguistic units.


\section*{Limitations}

\paragraph{Evaluation Criteria}

Although we have used coherence and consistency as evaluation criteria, compatible with the settings of  \citet{han2024rethinking, forde2024re}, we acknowledge that another common approach, based on SummEval \citep{fabbri2021summeval},  incorporates fluency, coherence, consistency, and relevance. However,  our previous experiments revealed an extremely low inter-annotator agreement rate (\textasciitilde{0}) on this schema, and suggested that annotators struggled to distinguish subtle differences between all four criteria. To mitigate this , we  focus on coherence and consistency, as they offer a more straightforward and reliable basis for evaluation.
We leave the question of how reliable are human and automatic metrics on those fine-grained aspects for future follow-up research.

\paragraph{Number of Annotations}
To cover diverse typological groups and resource levels while relying on available crowd workers, the number of annotations varies across languages. For example, Japanese had only one worker, leading to a smaller number of human annotations than other languages.

\section*{Acknowledgements}
This research was funded by a grant from the Israel Science Foundation (ISF) grant number 670/23 as well as  
a KAMIN grant from the Israeli Innovation Authority, for which we are grateful. We are further grateful for a generous VATAT grant to the BIU NLP team which contributed resources for computation, annotation and human valuation in this project. We further 
Omer Goldman and Uri Ernest, former PhD students at the BIU-NLP lab, for 
their kind consultation  on this work.

\bibliography{anthology}
\bibliographystyle{acl_natbib}

\appendix

\section{Data Collection}

\subsection{Language Selection}
\label{appendix:language_classes}
Table~\ref{tab:languages_classes} displays the full resource-type categorization per language we have defined using GPT-3 pre-trained data.

\subsection{Power Analysis for Sample Size}
\label{appendix:power_test}
To ensure the reliability of our statistical tests, we conducted a power analysis to determine the minimum required sample size for detecting a statistical correlation (\( p-value \leq 0.05 \)). we applied a t-test power analysis and computed the required sample size per group to achieve these conditions. The analysis revealed that a minimum of \textasciitilde{400} samples per language is necessary for a well-powered correlation.

\subsection{Participant Interface}
\label{appendix:ui}

\begin{figure*}
    \centering
    \scalebox{1}{
    \includegraphics[width=\textwidth]{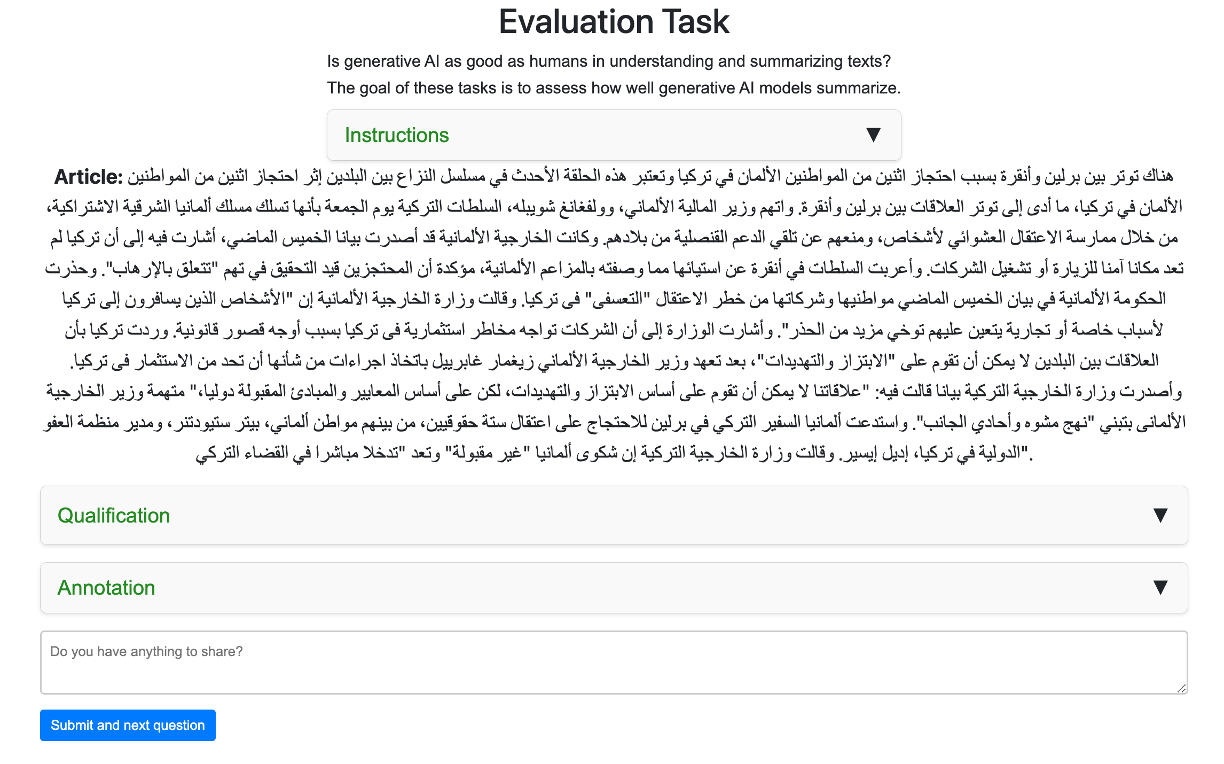}
    }
   \caption{\emph{Participant Interface in a closed mode:} The interface includes three drop-down sections: Instructions, Qualification and the Annotation task.}
    \label{fig:ui-all}
\end{figure*}

\begin{figure*}
    \centering
    \scalebox{1}{
    \includegraphics[width=\textwidth]{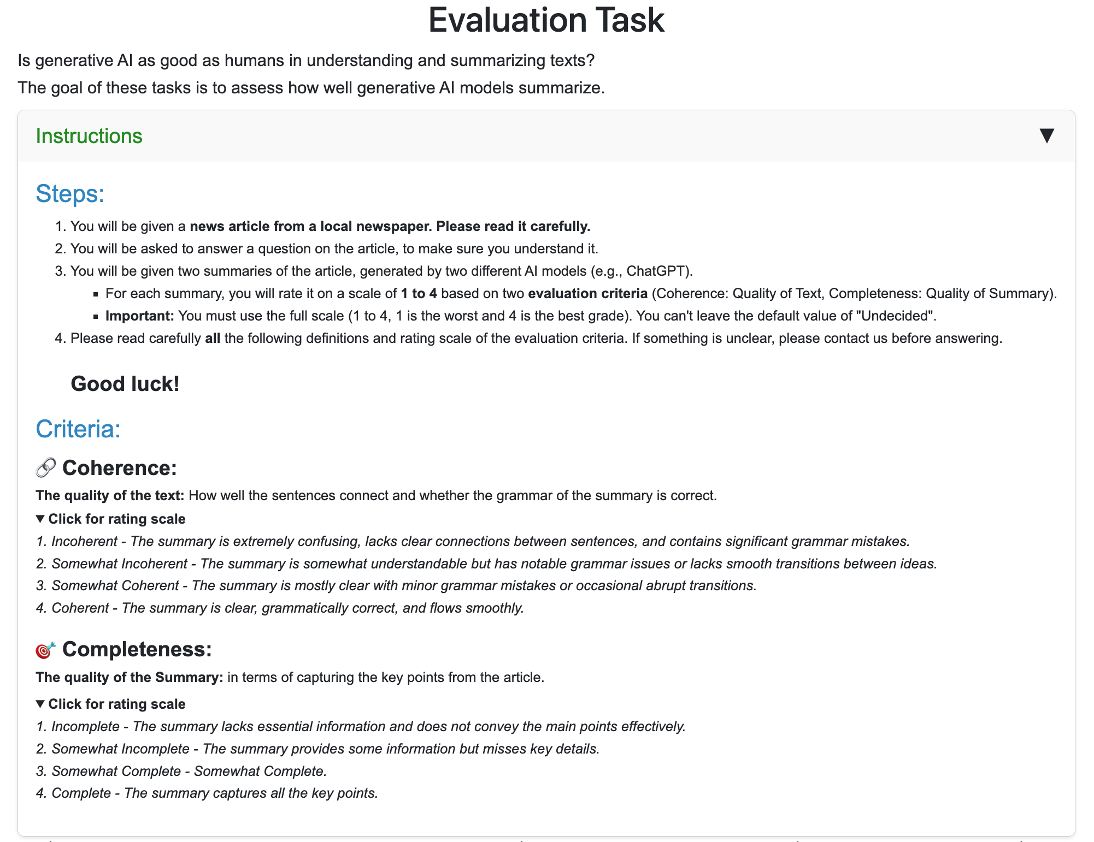}
    }
   \caption{\emph{The Participant Instructions Interface:}  The participant has general steps and a detailed explanation and examples of each tested criteria.}

    
    \label{fig:ui-instructions}
\end{figure*}

\begin{figure*}
    \centering
    \scalebox{1}{
    \includegraphics[width=\textwidth]{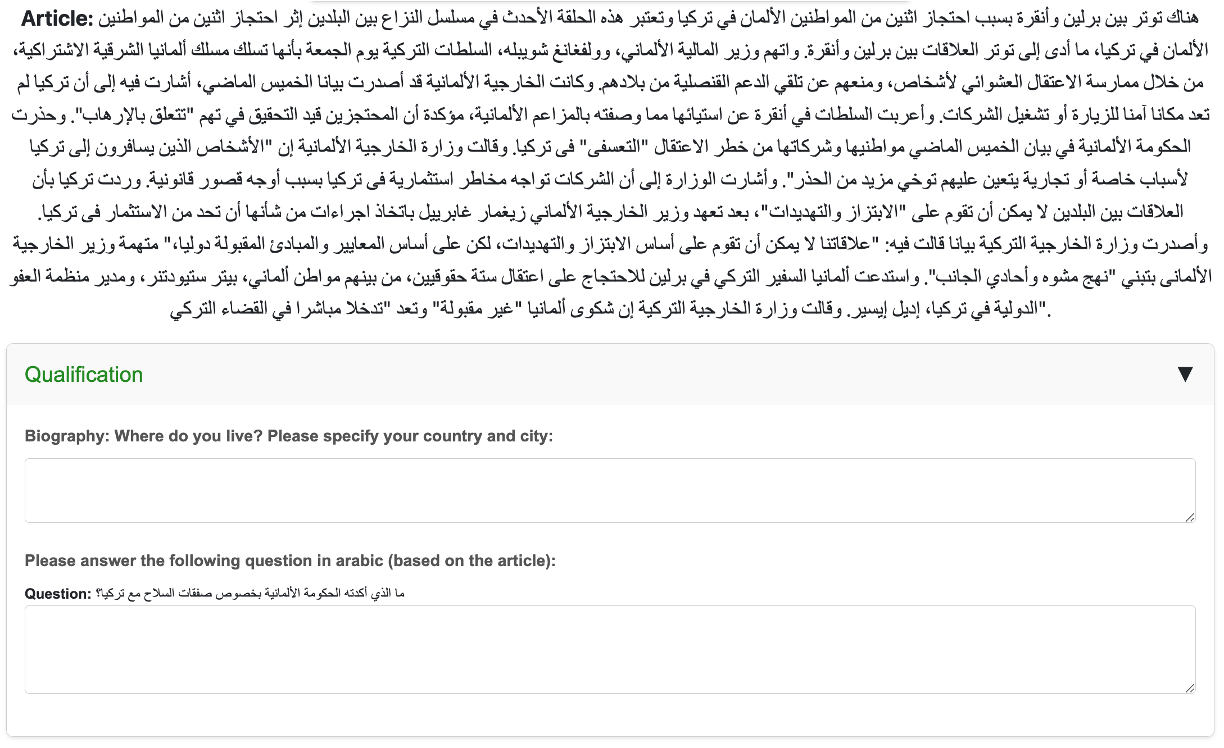}
    }
\caption{\emph{The Participant Qualification Interface:} The human summary is displayed at the top (the example is in Arabic), while the Qualification section below requires the participant to fill in their home state and answer a question generated by Gemini based on the human summary, designed to assess basic comprehension of the provided summary.}
\label{fig:ui-qualification}
\end{figure*}

\begin{figure*}
    \centering
    \scalebox{1}{
    \includegraphics[width=\textwidth]{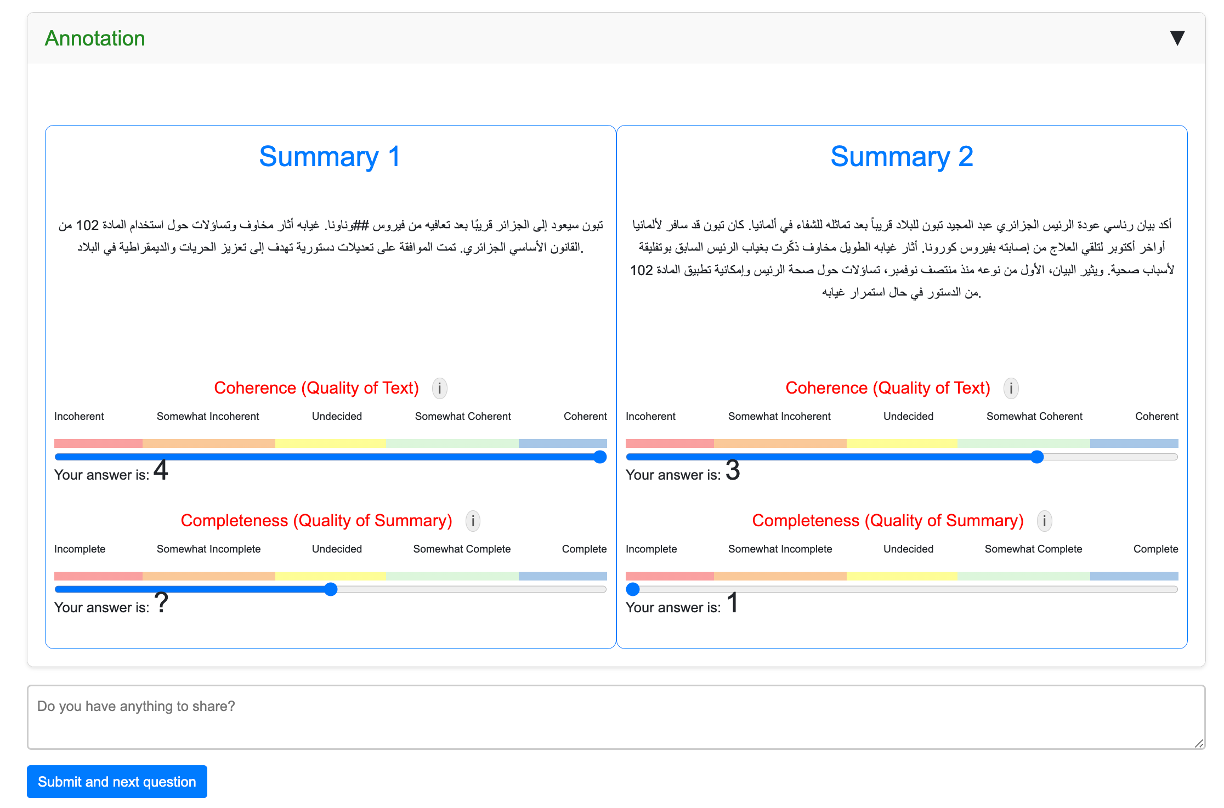}
    }
\caption{\emph{The Participant Annotation Interface:} Two summaries are displayed side by side. Each criterion includes a slider ranging from 1 to 4, along with an info hover feature providing a reminder of the criterion's definition.}
\label{fig:ui-annotation}
\end{figure*}

\begin{figure*}
    \centering
    \scalebox{1}{
    \includegraphics[width=\textwidth]{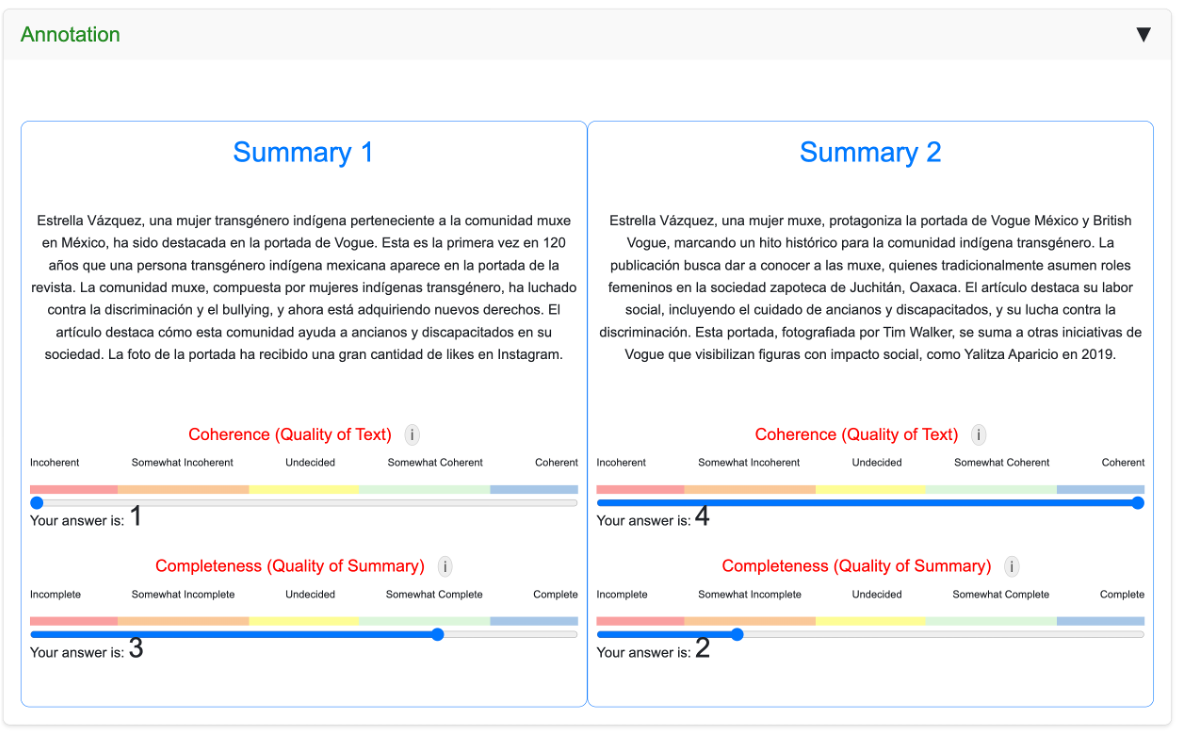}
    }
\caption{\emph{The Participant Annotation Interface:} displayed in Spanish}
\label{fig:ui-annotation_spanish}
\end{figure*}

The tasks are performed using a custom-built application displayed via mTurk, as shown in Figures \ref{fig:ui-all}-\ref{fig:ui-annotation}. The task is in Arabic, for example; see Figure \ref{fig:ui-annotation_spanish} for a Spanish example.

\subsection{Data Collection Details}
\label{appendix:protocol_details}

We utilized Amazon Mechanical Turk (MTurk) to distribute the task to various workers. For the student participants, all were undergraduate students from the linguistics field. To provide a custom user interface (UI) for our evaluation, we developed a JavaScript application and deployed it as a service using Google Cloud Run.\footnote{\url{https://cloud.google.com/run}}. Subsequently, we connected the MTurk participants to this service.

All participants were compensated in full, regardless of whether they correctly completed the task. The payment was set at \$2.5 for rating 5 pairs of summaries, which we estimated would take approximately 10--15 minutes to complete.

From lessoned learned from previous studies, we decided to invest significant effort into enhancing the user experience (UX) and the visual design of the application. This focus ensured that the interface was both intuitive and visually appealing, thereby improving participant engagement and task performance.

\subsection{Data Corruption}
\label{appendix:corruption}

We  experimented with the following corruption strategies on the generated summaries, addressing each of the quality criteria. \textbf{Coherence}: All verbs were replaced with their lemma forms, resulting in ungrammatical sentences. We removed random words from each sentence and replaced conjunctions with alternatives for languages without a lemmatizer (e.g., Chinese, Japanese, and Yoruba). 
In addition, we reorder pairs of sentences that are not adjacent. This corruption is inspired by the Shuffle Test \citet{barzilay2008modeling} used to evaluate whether models can detect incoherent text. \textbf{Completness}: 
Named entities with the same labels (e.g., PERSON and LOCATION) were shuffled within the summary. This is a common factual mistake of models \citep{pagnoni2021understanding}. Additionally, a random sentence from another article was inserted into the summary. Table \ref{tab:clean_corrupt_examples} provides an example for a clean sentence and it's corrupted version.

\begin{table*}[]
\centering
\scalebox{0.8}{
\begin{tabular}{lccc}
\toprule
\textbf{Language} & \textbf{BERT Model} & \textbf{NER Model} & \textbf{Lemmatizer} \\
\midrule
Turkish           & bert-base-turkish-cased & bert-base-turkish-cased-ner \footnote{\url{https://huggingface.co/CAMeL-Lab/bert-base-arabic-camelbert-msa-ner}} & zeyrek \footnote{\url{https://github.com/obulat/zeyrek}} \\
Hebrew            & DiktaBERT & DiktaBERT \citet{shmidman2023dictabert} & DiktaBERT \\
Arabic            & bert-base-arabic & CAMeL-Lab/bert-base-arabic-camelbert-msa-ner \footnote{\url{https://huggingface.co/CAMeL-Lab/bert-base-arabic-camelbert-msa-ner}} & qalsadi \footnote{\url{https://qalsadi.readthedocs.io/en/latest/README.html}} \\
Chinese           & bert-base-chinese & zh\_core\_web\_sm (spacy) & N.A \\
Japanese          & bert-base-japanese-v3 & ja\_core\_news\_sm (spacy) & N.A \\
Spanish           & bert-base-spanish-wwm-cased & es\_core\_news\_sm (spacy) & es\_core\_news\_md \\
Ukrainian         & bert-base-multilingual-cased & uk\_core\_news\_sm (spacy) & uk\_core\_news\_sm \\
Yoruba            & bert-base-multilingual-cased & N.A & N.A \\
\bottomrule
\end{tabular}
}
\caption{Language-specific BERT models, NER models, and lemmatizers.}
\label{tab:language_models_tokenizers}
\end{table*}

\begin{table*}[th]
\centering
\scalebox{0.8}{%
\begin{tabular}{@{}ccccc@{}}
\toprule
\textbf{Language}    & \textbf{Lang Code} & \textbf{Number of Tokens} & \textbf{Percentage of Tokens (\(p\%\))} & \textbf{Class}\\ \midrule
English     & en & 181,015 & 92.64\%  & A+             \\ \midrule
Spanish     & es &  \pz1,510   &  0.77289\% & A             \\
Japanese    & ja &  \pzz217   &  0.11109\% & A             \\
Chinese     & zh &  \pzz194   &  0.09905\% & A           \\  \midrule
Turkish     & tr &  \pzz116   &  0.05944\%  & B           \\ 
Arabic      & ar &  \pz\pz61  &  0.03114\% & A            \\
Hebrew      & he &  \pz\pz15  &  0.00769\% & B             \\
Ukrainian   & ukr &  \pz\pz14  &  0.00763\% & B             \\
Yoruba      & yor &  \pz\pz\pz0    &  0.00000\%  & B                   \\ \bottomrule
 \\ \hline
\end{tabular}%
}
\caption{ List of languages, language codes, number of tokens in pre-trained GPT-3 data, data ratios. The languages are grouped into two classes based on their data
ratios in the GPT-3 pre-trained data: High Resource
(\(p > 0.1\%\)), Low Resource (\(p < 0.1\%\))} 
\label{tab:languages_classes}
\end{table*}

\subsection{Qualification Task}
\label{appendix:qualification_task}

To filter out unqualified annotators, each was required to answer a generated question about the article in their native language. The model was prompted as follows:
{\ttfamily
Given the text: {<TEXT>} in {<LANGUAGE>}, generate a single-sentence question whose answer is found in the text.}

\section{Correlation Analysis}

\subsection{Implementation Details}
\label{appendix:multilingual_metrics}

\paragraph{Language-Specific BERT Models}
See Table \ref{tab:language_models_tokenizers} for the list of Bert models we used for each language. 

\paragraph{Python Libraries}
To use BERTScore (mBERT), we employed the official implementation.
For ROUGE (mBERT) and BPE tokenization, we used \textit{Multilingual-Rouge-Scorer}.\footnote{\url{https://github.com/faisaltareque/Multilingual-Rouge-Scorer/tree/main}}
For ROUGE (Language Tokenizer), we used the standard ROUGE package commonly applied in non-English papers.\footnote{\url{https://github.com/csebuetnlp/xl-sum/tree/master/multilingual_rouge_scoring}}
For other metrics, we used the implementation from SummEval \citep{fabbri2021summeval}.\footnote{\url{https://github.com/Yale-LILY/SummEval}}. We have used ChatGPT for assistance in coding the evaluation framework.

\subsection{Results}
\label{appendix:correlation_results_apendix}
See Table \ref{tab:correlations} for the full correlation for each language and metric. Also, Table \ref{tab:correlations_spearman} shows the correlation measured by 
Spearman's rank correlation coefficient.

\newpage

\begin{table*}[]
\centering
\scalebox{0.5}{
\begin{tabular}{lcccc}
\toprule
\textbf{Criterion}                    & \textbf{Rule}                  & \textbf{Example}                                                                                                                                                         \\ \hline
\multirow{3}{*}{\textbf{Coherence}}   & Replace with lemmas            & \textit{Clean:} The athletes are preparing for the championship. \newline \textit{Corrupt:} The athlete be prepare for the championship.                                 \\
                                      & Replace conjunctions           & \textit{Clean:} Policies address rising inflation. \newline \textit{Corrupt:} Policies however address rising inflation.                                                 \\ \cline{2-3} 
                                      & Reorder non-adjacent sentences & \textit{Clean:} The center is hosting a charity event. Volunteers are needed. \newline \textit{Corrupt:} Volunteers are needed. The center is hosting a charity event.   \\ \hline
\multirow{2}{*}{\textbf{Completness}} & Replace named entities         & \textit{Clean:} Joe Biden met Britney Spears at a charity event. \newline \textit{Corrupt:} Britney Spears, former president, met Joe Biden.                             \\ \cline{2-3} 
                                      & Insert irrelevant sentence     & \textit{Clean:} Scientists found a new fish species in the Amazon. \newline \textit{Corrupt:} Scientists found a new fish species. A bakery is giving free cake samples. \\ \cline{2-3} 
\end{tabular}%
}
\caption{Examples of clean and corrupt sentences based on coherence and completeness criteria.}
\label{tab:clean_corrupt_examples}
\end{table*}


\newpage

\begin{table*}[h]
\centering
\resizebox{\textwidth}{!}{%
\begin{tabular}{lcccccccccccccccc}
\toprule
                                & \multicolumn{8}{c}{\textbf{Coherence}}                                                                                                                                             & \multicolumn{8}{c}{\textbf{Completeness}}                                                                                                                                    \\ \cmidrule(lr){2-9} \cmidrule(lr){10-17} 
\textbf{Typological Family}     & \multicolumn{2}{c}{\textbf{Isolating}} & \multicolumn{2}{c}{\textbf{Agglutinative}} & \multicolumn{2}{c}{\textbf{High Fusional}} & \multicolumn{2}{c}{\textbf{Low Fusional}}      & \multicolumn{2}{c}{\textbf{Isolating}} & \multicolumn{2}{c}{\textbf{Agglutinative}} & \multicolumn{2}{c}{\textbf{High Fusional}} & \multicolumn{2}{c}{\textbf{Low Fusional}} \\
\textbf{Language Code}          & \textbf{ZH}       & \textbf{YOR}       & \textbf{JA}          & \textbf{TR}         & \textbf{AR}          & \textbf{HE}         & \textbf{ES} & \textbf{UKR}        & \textbf{ZH}       & \textbf{YOR}       & \textbf{JA}          & \textbf{TR}         & \textbf{AR}          & \textbf{HE}         & \textbf{ES}         & \textbf{UKR}        \\
\midrule
\rowcolor{gray!12} 
\multicolumn{17}{c}{\textbf{N-Gram Metrics}} \\ 
\midrule                                                                         
\makebox[0pt][r]{\scriptsize 1}\hspace{1em}\textbf{ROUGE1}                 & 0.06             & 0.06              & 0.25**               & 0.28*               & 0.14*                & -0.31**             & 0.16**      & 0.13*              & 0.11**            & 0.06             & 0.20*                & 0.08              & 0.19**               & -0.26**             & 0.11*               & 0.16*               \\
\makebox[0pt][r]{\scriptsize 2}\hspace{1em}\textbf{ROUGE2}                 & 0.07             & 0.08*              & 0.23*                & 0.31*               & 0.13*                & -0.14*              & 0.15*       & 0.08              & 0.12**            & 0.10             & 0.21*                & 0.13*               & 0.17*                & 0.06               & 0.10               & 0.08               \\
\makebox[0pt][r]{\scriptsize 3}\hspace{1em}\textbf{ROUGE3}                 & 0.08             & 0.06*               & 0.23*                & 0.28*               & 0.14*                & -0.07               & 0.08*       & 0.10*              & 0.12**            & 0.06*             & 0.19*                & 0.06                & 0.13*                & 0.18**               & 0.07                & -0.02               \\
\makebox[0pt][r]{\scriptsize 4}\hspace{1em}\textbf{ROUGEL}                 & 0.06             & 0.08             & 0.28*                & 0.28*               & 0.10*                & -0.26**             & 0.17**      & 0.09              & 0.10              & 0.10*             & 0.25*                & 0.09*               & 0.16*                & -0.26**             & 0.12*               & 0.13*              \\
\makebox[0pt][r]{\scriptsize 5}\hspace{1em}\textbf{CHRF}                   & 0.08             & 0.02               & 0.27*                & 0.25*               & 0.12*                & -0.21**             & 0.15**      & 0.17**              & 0.10*             & 0.02            & 0.23**                & 0.19*               & 0.18*                & -0.41**              & 0.13*               & 0.17**              \\
\makebox[0pt][r]{\scriptsize 6}\hspace{1em}\textbf{BLEU}                   & 0.08             & 0.10*              & N.A                  & 0.24*               & 0.14*                & -0.16*              & 0.11**      & 0.15*              & -0.05             & 0.11*            & 0.24**                 & 0.05                & 0.12*                & -0.38**               & 0.06               & 0.04               \\ \hdashline
\makebox[0pt][r]{\scriptsize 7}\hspace{1em}\textbf{ROUGEL (mBERT Tokenizer)} & 0.10**            & 0.07*              & 0.13*                & 0.21*               & 0.03                & 0.36**              & 0.13**      & 0.04             & 0.08             & 0.09*              & 0.09*                & 0.03                & 0.12*                & 0.40**              & 0.09*               & 0.12*                \\
\makebox[0pt][r]{\scriptsize 8}\hspace{1em}\textbf{ROUGEL (Language Tokenizer)} & 0.07           & -0.02            & 0.10*                & 0.20*               & 0.04                & 0.30*                & 0.13*       & 0.11*              & 0.04             & -0.02               & 0.12*                & 0.06                & 0.17*                & 0.40**              & 0.11*                & 0.12*                \\
\midrule
\rowcolor{gray!12} 
\multicolumn{17}{c}{\textbf{Neural-Based Metrics}} \\ 
\midrule
\makebox[0pt][r]{\scriptsize 11}\hspace{1em}\textbf{BERTScore Monolingual}  & 0.10*             & -0.02               & 0.30                 & 0.33*               & 0.10*                & 0.0                 & 0.24**      & 0.12*              & 0.16              & 0.01               & 0.26*                & 0.11**              & 0.14*                & 0.13                & 0.15*               & 0.21**              \\
\makebox[0pt][r]{\scriptsize 12}\hspace{1em}\textbf{BERTScore (mBERT)} & 0.12*             & 0.02              & 0.27*                & 0.25*               & 0.08*                & -0.15*              & 0.22**      & 0.11*              & 0.21              & 0.03              & 0.24*                & 0.10*               & 0.12*                & -0.06               & 0.15*               & 0.15*              \\
\makebox[0pt][r]{\scriptsize 13}\hspace{1em}\textbf{COMET}                  & 0.13*             & 0.00               & 0.27*                & 0.23*               & 0.00                 & 0.38                & 0.27**      & 0.16              & 0.23**            & 0.02            & 0.24*                & 0.11*               & 0.25**              & 0.49**              & 0.09               & 0.25*               \\
\makebox[0pt][r]{\scriptsize 14}\hspace{1em}\textbf{Gemini Model}           & 0.07*            & 0.11*               & 0.27                & 0.08*               & 0.03                 & -0.10               & 0.16**      & 0.16**             & 0.05             & 0.16*             & 0.23*                & 0.19**              & 0.19**               & 0.12                & 0.06                & 0.06               \\ \midrule
\end{tabular}%
}
\caption{Spearman correlation between language and evaluation metrics. Significance levels are denoted by: * \(p < 0.05\), ** \(p < 0.01\). The dashed line separates the English-based metrics from the multilingual metrics.}
\label{tab:correlations_spearman}
\end{table*}

\newpage

\begin{table*}[]
\centering
\resizebox{\textwidth}{!}{%
\begin{tabular}{lcccccccccccccccc}
\toprule
                                & \multicolumn{8}{c}{\textbf{Coherence}}                                                                                                                                             & \multicolumn{8}{c}{\textbf{Completeness}}                                                                                                                                    \\ \cmidrule(lr){2-9} \cmidrule(lr){10-17} 
\textbf{Typological Family}     & \multicolumn{2}{c}{\textbf{Isolating}} & \multicolumn{2}{c}{\textbf{Agglutinative}} & \multicolumn{2}{c}{\textbf{High Fusional}} & \multicolumn{2}{c}{\textbf{Low Fusional}}      & \multicolumn{2}{c}{\textbf{Isolating}} & \multicolumn{2}{c}{\textbf{Agglutinative}} & \multicolumn{2}{c}{\textbf{High Fusional}} & \multicolumn{2}{c}{\textbf{Low Fusional}} \\
\textbf{Language Code}          & \textbf{ZH}       & \textbf{YOR}       & \textbf{JA}          & \textbf{TR}         & \textbf{AR}          & \textbf{HE}         & \textbf{ES} & \textbf{UKR}        & \textbf{ZH}       & \textbf{YOR}       & \textbf{JA}          & \textbf{TR}         & \textbf{AR}          & \textbf{HE}         & \textbf{ES}         & \textbf{UKR}        \\
\midrule
\rowcolor{gray!12} 
\multicolumn{17}{c}{\textbf{N-Gram Metrics}} \\ 
\midrule                                                                         
\makebox[0pt][r]{\scriptsize 1}\hspace{1em}\textbf{ROUGE1}                 & 0.09*             & 0.12*              & 0.25**               & 0.33*               & 0.10*                & -0.31**             & 0.18**      & 0.09*              & 0.10**            & 0.11**             & 0.15*                & 0.08**              & 0.14**               & -0.23**             & 0.13*               & 0.15*               \\
\makebox[0pt][r]{\scriptsize 2}\hspace{1em}\textbf{ROUGE2}                 & 0.11*             & 0.14*              & 0.20*                & 0.45*               & 0.10*                & -0.14*              & 0.14*       & 0.06*              & 0.10**            & 0.11**             & 0.16*                & 0.12*               & 0.13*                & -0.06               & 0.11*               & 0.10*               \\
\makebox[0pt][r]{\scriptsize 3}\hspace{1em}\textbf{ROUGE3}                 & 0.11*             & 0.08               & 0.20*                & 0.36*               & 0.12*                & -0.07               & 0.08*       & 0.08*              & 0.10**            & 0.08**             & 0.16*                & 0.07                & 0.11*                & -0.01               & 0.07                & 0.00                \\
\makebox[0pt][r]{\scriptsize 4}\hspace{1em}\textbf{ROUGEL}                 & 0.10*             & 0.14**             & 0.23*                & 0.34*               & 0.06*                & -0.26**             & 0.19**      & 0.06*              & 0.10              & 0.12**             & 0.19*                & 0.09*               & 0.10*                & -0.21**             & 0.13*               & 0.11**              \\
\makebox[0pt][r]{\scriptsize 5}\hspace{1em}\textbf{CHRF}                   & 0.10*             & 0.11               & 0.22*                & 0.31*               & 0.09*                & -0.21**             & 0.18**      & 0.12*              & 0.10*             & -0.16**            & 0.18*                & 0.11*               & 0.14*                & -0.12*              & 0.14*               & 0.15**              \\
\makebox[0pt][r]{\scriptsize 6}\hspace{1em}\textbf{BLEU}                   & -0.03             & 0.13*              & N.A                  & 0.36*               & 0.06*                & -0.16*              & 0.10**      & 0.10*              & -0.03             & -0.38**            & 0.10                 & 0.05                & 0.09(                & -0.08               & 0.10*               & 0.00*               \\ \hdashline
\makebox[0pt][r]{\scriptsize 7}\hspace{1em}\textbf{ROUGEL (mBERT Tokenizer)} & -0.05*            & 0.10*              & 0.30*                & 0.28*               & 0.09*                & 0.46**              & 0.18**      & 0.12*              & 0.11*             & 0.10*              & 0.21*                & 0.05                & 0.12*                & 0.49**              & 0.09                & 0.10                \\
\makebox[0pt][r]{\scriptsize 8}\hspace{1em}\textbf{ROUGEL (Language Tokenizer)} & -0.06*            & 0.14**             & 0.25*                & 0.32*               & 0.11*                & 0.3*                & 0.17*       & 0.10*              & 0.08*             & 0.00               & 0.21*                & 0.08                & 0.17*                & 0.47**              & 0.09                & 0.09                \\
\makebox[0pt][r]{\scriptsize 9}\hspace{1em}\textbf{ROUGEL (Llema Form)}    & N.A               & N.A                & N.A                  & 0.29*               & 0.09                 & 0.42*               & 0.15*       & 0.14**             & N.A               & N.A                & N.A                  & 0.10*               & 0.12*                & 0.46*               & 0.10                & 0.09*               \\ 
\midrule
\rowcolor{gray!12} 
\multicolumn{17}{c}{\textbf{Neural-Based Metrics}} \\ 
\midrule
\makebox[0pt][r]{\scriptsize 10}\hspace{1em}\textbf{BERTScore}              & 0.02              & 0.16*              & 0.16*                & 0.07                & 0.10*                & -0.01               & 0.19**      & 0.09*              & 0.20**            & 0.25**             & 0.17*                & 0.12*               & 0.14*                & 0.17*               & 0.17*               & 0.13*               \\ \hdashline
\makebox[0pt][r]{\scriptsize 11}\hspace{1em}\textbf{BERTScore Monolingual}  & 0.12*             & 0.04               & 0.09                 & 0.31*               & 0.09*                & 0.0                 & 0.27**      & 0.07*              & 0.09              & 0.07               & 0.19*                & 0.11**              & 0.15*                & 0.11                & 0.17*               & 0.15**              \\
\makebox[0pt][r]{\scriptsize 12}\hspace{1em}\textbf{BERTScore (mBERT)} & 0.11*             & 0.09*              & 0.22*                & 0.23*               & 0.05*                & -0.15*              & 0.23**      & 0.08*              & 0.10              & 0.15*              & 0.17*                & 0.09*               & 0.09*                & -0.07               & 0.16*               & 0.21**              \\
\makebox[0pt][r]{\scriptsize 13}\hspace{1em}\textbf{COMET}                  & 0.08*             & 0.01               & 0.21*                & 0.21*               & 0.10                 & 0.38                & 0.32**      & 0.11*              & 0.24**            & 0.06**            & 0.18*                & 0.09*               & 0.23**              & 0.17**              & 0.14*               & 0.25*               \\
\makebox[0pt][r]{\scriptsize 14}\hspace{1em}\textbf{Gemini Model}           & 0.16**            & 0.01               & 0.23*                & 0.08*               & 0.03                 & -0.10               & 0.19**      & 0.16**             & 0.11*             & 0.16**             & 0.20*                & 0.16**              & 0.16**               & 0.00                & 0.09                & 0.10*               \\ \midrule
\end{tabular}%
}
\caption{Pearson correlation between language and evaluation metrics. Significance levels are denoted by: * \(p < 0.05\), ** \(p < 0.01\). The dashed line separates the English-based metrics from the multilingual metrics.}
\label{tab:correlations}
\end{table*}

\end{document}